\documentclass{article}
\usepackage{booktabs}       
\usepackage{amsfonts}       
\usepackage{microtype}      
\usepackage{xcolor}
\usepackage{url}
\usepackage{verbatim} 
\usepackage{graphicx}
\usepackage{caption} 
\usepackage{multirow}
\usepackage{xspace}
\usepackage{epsfig}
\usepackage{amsmath}
\usepackage{amsthm}
\usepackage{amssymb}
\usepackage{times}
\usepackage{xr}
\usepackage{bbm}
\usepackage{dsfont}
\usepackage{bm}
\usepackage{subcaption}
\usepackage{enumitem}
\usepackage{hyperref}       

\newcommand{\xhdr}[1]{{\noindent\bfseries #1}.}

\newcommand{\CITE}{{\textcolor{red}{[CITE]}}}
\newcommand{\name}{P-GNN\xspace}

\newcommand{\mb}{\mathbf}
\newcommand{\cut}[1]{}

\newtheorem{proposition}{Proposition}
\newtheorem{theorem}{Theorem}

\newtheorem{definition}{Definition}

\usepackage[accepted]{icml2019}

\icmltitlerunning{Position-aware Graph Neural Networks}

\begin{document}

\twocolumn[
\icmltitle{Position-aware Graph Neural Networks}



\icmlsetsymbol{equal}{*}

\begin{icmlauthorlist}
\icmlauthor{Jiaxuan You}{stanford}
\icmlauthor{Rex Ying}{stanford}
\icmlauthor{Jure Leskovec}{stanford}
\end{icmlauthorlist}

\icmlaffiliation{stanford}{Department of Computer Science, Stanford University, Stanford, CA, USA}

\icmlcorrespondingauthor{Jiaxuan You}{jiaxuan@cs.stanford.edu}
\icmlcorrespondingauthor{Jure Leskovec}{jure@cs.stanford.edu}
\icmlkeywords{Machine Learning, ICML}

\vskip 0.3in
]



\printAffiliationsAndNotice{}  

\begin{abstract}
 Learning node embeddings that capture a node's position within the broader graph structure is crucial for many prediction tasks on graphs.
However, existing Graph Neural Network (GNN) architectures have limited power in capturing the position/location of a given node with respect to all other nodes of the graph.
Here we propose {\em Position-aware Graph Neural Networks (P-GNNs)}, a new class of GNNs for computing position-aware node embeddings. P-GNN first samples sets of anchor nodes, computes the distance of a given target node to each anchor-set, and then learns a non-linear distance-weighted aggregation scheme over the anchor-sets. This way P-GNNs can capture positions/locations of nodes with respect to the anchor nodes.
P-GNNs have several advantages: they are inductive, scalable, and can incorporate node feature information.
We apply P-GNNs to multiple prediction tasks including link prediction and community detection. We show that P-GNNs consistently outperform state of the art GNNs, with up to 66\% improvement in terms of the ROC AUC score.

\cut{
Learning node embeddings that are aware of the position of nodes within a graph is crucial for many prediction tasks such as node classification, link prediction, and community detection.
However, while expressive and most popular, existing Graph Neural Network (GNN) approaches have limited power for representing positions/locations of nodes in a bigger network structure.
Here we propose {\em Position-aware Graph Neural Networks (P-GNN)}, a new class of GNNs for computing position-aware node embeddings and remains inductive, via computing shortest path based distances between a node and randomly selected node subsets.
We apply P-GNN to multiple prediction tasks including link prediction and community detection. We show that both variants of P-GNN consistently outperforms state of the art GNN variants by 20\% in transductive settings, while significantly outperforms GNN models in inductive settings by 50\%. 
}
\end{abstract}

 \section{Introduction}

Learning low-dimensional vector representations of nodes in graphs \cite{hamilton2017representation} has led to advances on tasks such as node classification \cite{kipf2016semi}, link prediction \cite{grover2016node2vec}, graph classification \cite{ying2018hierarchical} and graph generation \cite{you2018graphrnn}, with successful applications across domains such as social and information networks \cite{ying2018graph}, chemistry \cite{you2018graph}, and biology \cite{zitnik2017predicting}.

Node embedding methods can be categorized into Graph Neural Networks (GNNs) approaches \cite{scarselli2009graph}, matrix-factorization approaches \cite{belkin2002laplacian}, and random-walk approaches \cite{perozzi2014deepwalk}. Among these, GNNs are currently the most popular paradigm, largely owing to their efficiency and inductive learning capability~\cite{hamilton2017inductive}. By contrast, random-walk approaches~\cite{perozzi2014deepwalk,grover2016node2vec} are limited to transductive settings and cannot incorporate node attributes. In the GNN framework, the embedding of a node is computed by a GNN layer aggregating information from the node's network neighbors via non-linear transformation and aggregation functions~\cite{battaglia2018relational}. Long-range node dependencies can be captured via stacking multiple GNN layers, allowing the information to propagate for multiple-hops~\cite{xu2018representation}. 

\begin{figure}
    \centering
    \includegraphics[width=0.42\textwidth]{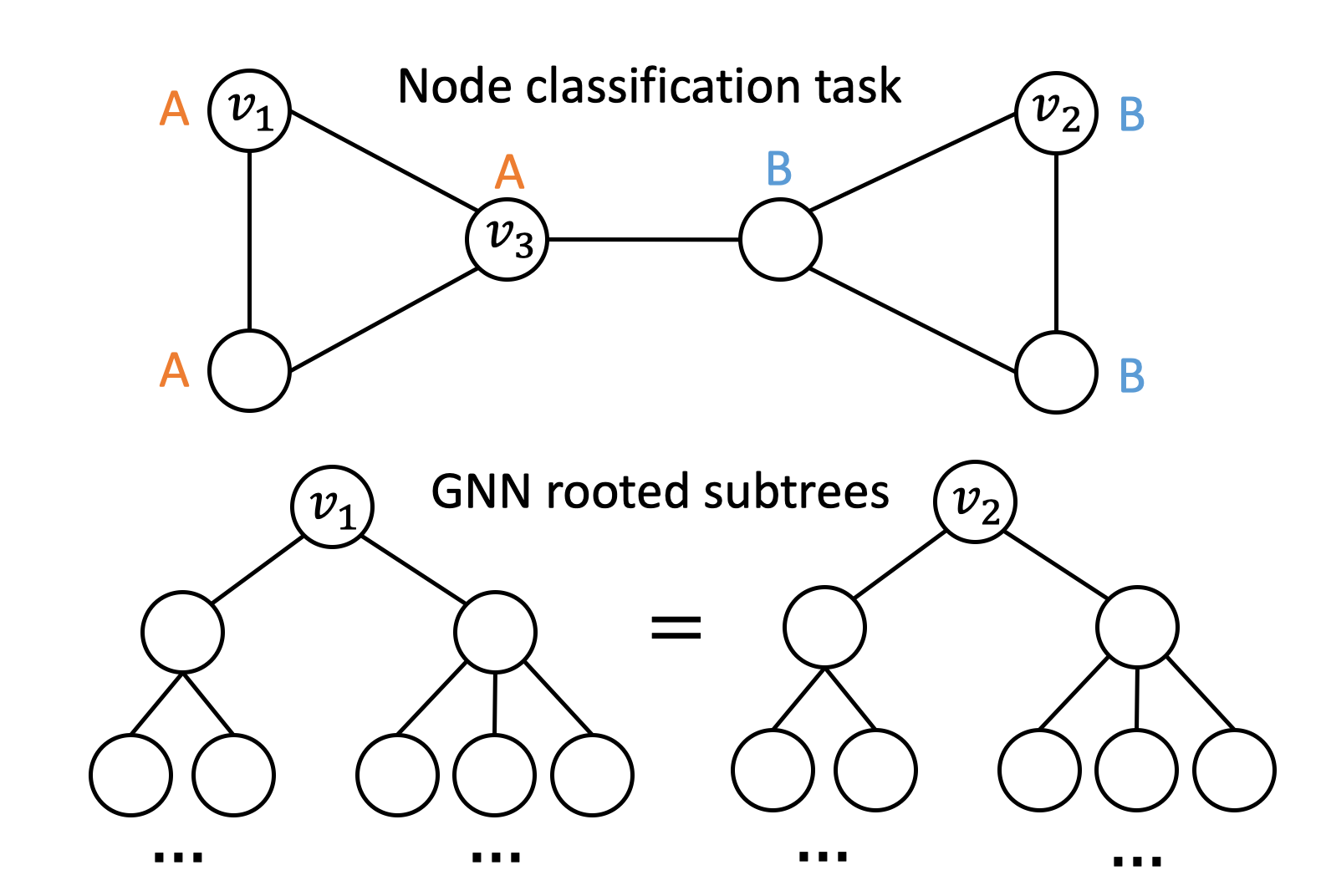}
    \caption{Example graph where GNN is not able to distinguish and thus classify nodes $v_1$ and $v_2$ into different classes based on the network structure alone. (Note we do not consider node features.)
    Each node is labeled based on its label $A$ or $B$, and effective node embedding should be able to learn to distinguish nodes $v_1$ and $v_2$ (that is, embed them into different points in the space). However, GNNs, regardless of depth, will \textit{always} assign the same embedding to both nodes,
    because the two nodes are symmetric/isomorphic in the graph, and their GNN rooted subtrees used for message aggregation are the same. In contrast, P-GNNs can break the symmetry by using $v_3$ as the anchor-set, then the shortest path distances $(v_1, v_3)$ and $(v_2, v_3)$ are different and nodes $v_1$ and $v_2$ can thus be distinguished.
    }
    \label{fig:example}
\vspace{-3mm}
\end{figure}

However, the key limitation of existing GNN architectures is that they fail to capture the {\em position/location} of the node within the broader context of the graph structure. 
For example, if two nodes reside in very different parts of the graph but have topologically the same (local) neighbourhood structure, they will have identical GNN structure. Therefore, the GNN will embed them to the same point in the embedding space (we ignore node attributes for now). Figure~\ref{fig:example} gives an example where a GNN cannot distinguish between nodes $v_1$ and $v_2$ and will always embed them to the same point because they have isomorphic network neighborhoods. Thus, GNNs will never be able to classify nodes $v_1$ and $v_2$ into different classes because from the GNN point of view they are indistinguishable (again, not considering node attributes).
%
Researchers have spotted this weakness~\cite{xu2018powerful} and developed heuristics to fix the issue: augmenting node features with one-hot encodings \cite{kipf2016semi}, or making GNNs deeper~\cite{selsam2018learning}. However, models trained with one-hot encodings cannot generalize to unseen graphs, and arbitrarily deep GNNs still cannot distinguish structurally isomorphic nodes (Figure \ref{fig:example}).

Here we propose {\em Position-aware Graph Neural Networks (P-GNNs)}, a new class of Graph Neural Networks for computing node embeddings that incorporate a node's positional information with respect to all other nodes in the network, while also retaining inductive capability and utilizing node features.
Our key observation is that node position can be captured by a low-distortion embedding by quantifying the distance between a given node and a set of anchor nodes.
Specifically, P-GNN uses a sampling strategy with theoretical guarantees to choose $k$ random subsets of nodes called {\em anchor-sets}. 
To compute a node's embedding, P-GNN first samples multiple anchor-sets in each forward pass, then learns a non-linear aggregation scheme that combines node feature information from each anchor-set and weighs it by the distance between the node and the anchor-set. Such aggregations can be naturally chained and combined into multiple layers to enhance model expressiveness.
Bourgain theorem \cite{bourgain1985lipschitz} guarantees that only $k = O(\log^2 n)$ anchor-sets are needed to preserve the distances in the original graph with low distortion. 

We demonstrate the P-GNN framework in various real-world graph-based prediction tasks. In settings where node attributes are not available, P-GNN's computation of the $k$ dimensional distance vector is inductive across different node orderings and different graphs.
When node attributes are available, a node's embedding is further enriched by aggregating information from all anchor-sets, weighted by the $k$ dimensional distance vector.
Furthermore, we show theoretically that P-GNNs are more general and expressive than traditional message-passing GNNs. In fact, message-passing GNNs can be viewed as special cases of P-GNNs with degenerated distance metrics and anchor-set sampling strategies.
In large-scale applications, computing distances between nodes can be prohibitively expensive. Therefore, we also propose P-GNN-Fast which adopts approximate node distance computation. We show that P-GNN-Fast has the same computational complexity as traditional GNN models while still preserving the benefits of P-GNN.

We apply P-GNNs to 8 different datasets and several different prediction tasks including link prediction and community detection\footnote{Code and data are available in \url{https://github.com/JiaxuanYou/P-GNN/}}. In all datasets and prediction tasks, we show that P-GNNs consistently outperforms state of the art GNN variants, with up to 66\% AUC score improvement.

 \section{Related Work}

Existing GNN models belong to a family of graph message-passing architectures that use different aggregation schemes for a node to aggregate feature messages from its neighbors in the graph: Graph Convolutional Networks use mean pooling \cite{kipf2016semi}; GraphSAGE concatenates the node's feature in addition to mean/max/LSTM pooled neighborhood information \cite{hamilton2017inductive}; Graph Attention Networks aggregate neighborhood information according to trainable attention weights \cite{velickovic2017graph}; Message Passing Neural Networks further incorporate edge information when doing the aggregation \cite{gilmer2017neural}; And, Graph Networks further consider global graph information during aggregation \cite{battaglia2018relational}.
However, all these models focus on learning node embeddings that capture local network structure around a given node. Such models are at most as powerful as the WL graph isomorphism test \cite{xu2018powerful}, which means that they cannot distinguish nodes at symmetric/isomorphic positions in the network (Figure~\ref{fig:example}). That is, without relying on the node feature information, above models will always embed nodes at symmetric positions into same embedding vectors, which means that such nodes are indistinguishable from the GNN's point of view.


Heuristics that alleviate the above issues include assigning an unique identifier to each node \cite{kipf2016semi,hamilton2017inductive} or using locally assigned node identifiers plus pre-trained transductive node features \cite{zhang2018link}.
However, such models are not scalable and cannot generalize to unseen graphs where the canonical node ordering is not available. In contrast, P-GNNs can capture positional information without sacrificing other advantages of GNNs.

One alternative method to incorporate positional information is utilizing a graph kernel,
which crucially rely on the positional information of nodes and inspired our P-GNN model.
Graph kernels implicitly or explicitly map graphs to a Hilbert space. 
Weisfeiler-Lehman and Subgraph kernels have
been incorporated into deep graph kernels~\cite{Yan+2015} to capture structural properties of neighborhoods.
\citeauthor{Gae+2003} (\citeyear{Gae+2003})
and \citeauthor{Kas+2003} (\citeyear{Kas+2003}) also proposed
graph kernels based on random walks, which count the number of walks
two graphs have in common~\cite{Sug+2015}.
Kernels based on shortest paths were first proposed in~\cite{Borgwardt2005}.


\section{Preliminaries}

\begin{figure*}
    \centering
    \includegraphics[width=\textwidth]{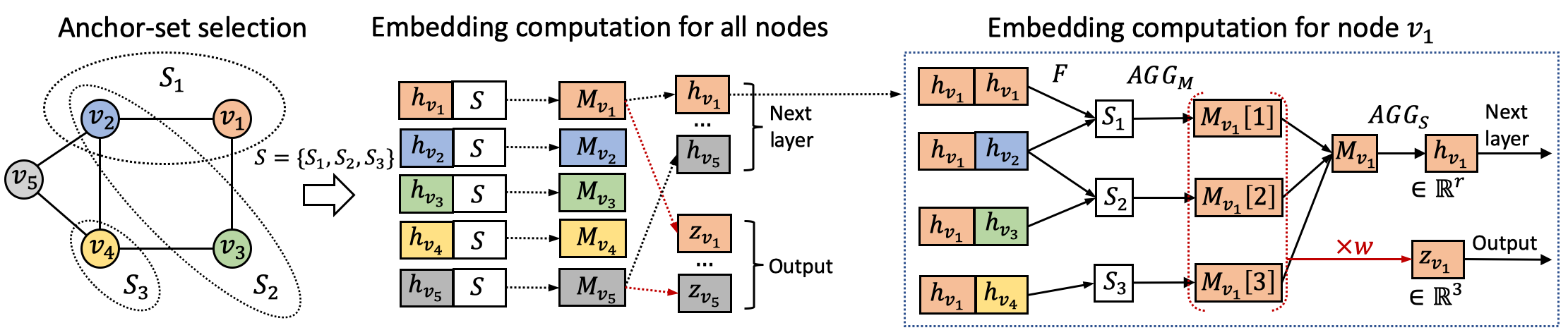}
    \caption{P-GNN architecture. P-GNN first samples multiple anchor-sets $S = \{S_1, S_2, S_3\}$ of different sizes (\textbf{Left}). Then, position-aware node embeddings $\mb{z}_{v_i}$ are computed via messages $M_{v_i}$ between a given node $v_i$ and the anchor-sets $S_i$ which are shared across all the nodes (\textbf{Middle}). To compute the embedding $\mb{z}_{v_1}$ for node $v_1$, one layer of P-GNN first computes messages via function $F$ and then aggregates them via a learnable function $\textsc{Agg}_M$ over the nodes in each anchor-set $S_i$ to obtain a matrix of anchor-set messages $\mb{M}_{v_1}$. The message matrix $\mb{M}_{v_1}$ is then further aggregated using a learnable function $\textsc{Agg}_S$ to obtain node $v_1$'s message $\mb{h}_{v_1}$ that can be passed to the next level of P-GNN. At the same time a learned vector $\mb{w}$ is used reduce $\mb{M}_{v_1}$ into a fixed-size position-aware embedding $\mb{z}_{v_1}$ which is the output of the P-GNN (\textbf{Right}).
    }
    \label{fig:PGNN}
\end{figure*}

\subsection{Notation and Problem Definition}
A graph can be represented as $G = (\mathcal{V},\mathcal{E})$, where $\mathcal{V} = \{v_1, ..., v_n\}$ is the node set and $\mathcal{E}$ is the edge set. In many applications where nodes have attributes, we augment $G$ with the node feature set $\mathcal{X} = \{\mb{x}_1, ..., \mb{x}_n\}$ where $\mb{x}_i$ is the feature vector associated with node $v_i$. 

Predictions on graphs are made by first embedding nodes into a low-dimensional space which is then fed into a classifier, potentially in an end-to-end fashion. %
Specifically, a node embedding model can be written as a function $f: \mathcal{V} \rightarrow \mathcal{Z}$ that maps nodes $\mathcal{V}$ to $d$-dimensional vectors $\mathcal{Z} = \{\mb{z}_1, ..., \mb{z}_n\}, \mb{z}_i \in \mathbb{R}^d$.


\subsection{Limitations of Structure-aware Embeddings}
\label{sc:limitation_structure_aware}
\cut{
We identify two key properties of node embeddings that are of special interests in applications, which we name as \textit{structure-aware} embeddings and \textit{position-aware} embeddings.
Structure-aware embeddings detect the structural role for a given node, usually involve some approximate graph isomorphism test. 
Position-aware embeddings reflect homophily properties in the graph \cite{grover2016node2vec}, which is highly related to the shortest path distance metrics.
Usually, tasks such as community detection and link prediction require more of position-aware embeddings, while common node classification tasks reflect more of structure-aware distances \CITE.
Although both types of embeddings have been used to qualitatively discuss the nature of various network prediction tasks, their relationship is not clearly understood. 
}

Our goal is to learn embeddings that capture the local network structure as well as retain the global network position of a given node.
We call node embeddings to be {\em position-aware}, if the embedding of two nodes can be used to (approximately) recover their shortest path distance in the network. This property is crucial for many prediction tasks, such as link prediction and community detection.
We show below that GNN-based embeddings cannot recover shortest path distances between nodes, which may lead to suboptimal performance in tasks where such information is needed.

\begin{definition}
A node embedding $\mb{z}_i = f_p(v_i), \forall v_i \in \mathcal{V}$ is position-aware if there exists a function $g_p(\cdot, \cdot)$ such that $d_{sp}(v_i, v_j) = g_p(\mb{z}_i, \mb{z}_j)$, where $d_{sp}(\cdot, \cdot)$ is the shortest path distance in $G$.
\end{definition}

\begin{definition}
A node embedding $\mb{z}_i = f_{s_q}(v_i), \forall v_i \in \mathcal{V}$ is structure-aware if it is a function of up to $q$-hop network neighbourhood of node $v_i$. Specifically, $\mb{z}_i = g_s(N_1(v_i),...,N_q(v_i))$, where $N_k(v_i)$ is the set of the nodes $k$-hops away from node $v_i$, and $g_s$ can be any function.
\end{definition}

For example, most graph neural networks compute node embeddings by aggregating information from each node's $q$-hop neighborhood, and are thus structure-aware. In contrast, (long) random-walk-based embeddings such as DeepWalk and Node2Vec are position-aware, since their objective function forces nodes that are close in the shortest path to also be close in the embedding space. 
In general, structure-aware embeddings cannot be mapped to position-aware embeddings.
Therefore, when the learning task requires node positional information, only using structure-aware embeddings as input is not sufficient:

\begin{proposition}
\label{prop:emb_dichoto}
There exists a mapping $g$ that maps structure-aware embeddings $f_{s_q}(v_i), \forall v_i \in \mathcal{V}$ to position-aware embeddings $f_p(v_i), \forall v_i \in \mathcal{V}$, if and only if no pair of nodes have isomorphic local $q$-hop neighbourhood graphs.
\end{proposition}

Proposition \ref{prop:emb_dichoto} is proved in the Appendix. The proof is based on the identifiability arguments similar to the proof of Theorem 1 in \cite{hamilton2017inductive}, and also explains why in some cases GNNs may perform well in tasks requiring positional information. However, in real-world graphs such as molecules and social networks, the structural equivalences between nodes' local neighbourhood graphs are quite common, making GNNs hard to identify different nodes.
Furthermore, the mapping $g$ essentially memorizes the shortest path distance between a pair of structure-aware node embeddings whose local neighbourhoods are unique. Therefore, even if the GNN perfectly learns the mapping $g$, it cannot generalize to the mapping to new graphs.

\cut{
We concretely state the limitation of structure-aware node embeddings in the inductive setting in Proposition \ref{prop:emb_dichoto_inductive} (proof in the Appendix).

\begin{proposition}
\label{prop:emb_dichoto_inductive}
There does not exist a general mapping $g$ that maps from structure-aware embeddings $f_{s_q}(v_i)$ to shortest path distance $d_{sp}$ for any given graph.
\end{proposition}
}


\cut{
Finally, we point out that while inductive position-aware link-prediction tasks are well-defined and prevalent, inductive position-aware node classification tasks are in fact ill-defined\footnote{Note that position-aware node classification tasks well defined in the transductive setting}.
For different 

In other words, the predictions for position-aware node classification tasks only need to match the node labels \emph{up to permutation of labels}, which is equivalent to link prediction where the links represents the equivalence relation between nodes, and the labels form the quotient set of $V$.

\begin{proposition}
\label{prop:node_transform_edge}
Inductive position-aware node labels are in fact ill-defined. 
There does not exist a position-aware node labeling function for a dataset of multiple graphs, such that the labeling function remains consistent for each input node after permutations of the graphs.
\end{proposition} 
}

\section{Proposed Approach}

In this section, we first describe the \name framework that extends GNNs to learn position-aware node embeddings. We follow by a discussion on our model designing choices. Last, we theoretically show how P-GNNs generalize existing GNNs and learn position-aware embeddings.

\subsection{The Framework of P-GNNs}

We propose Position-aware Graph Neural Networks that generalize the concepts of Graph Neural Networks with two key insights.
First, when computing the node embedding, instead of only aggregating messages computed from a node's local network neighbourhood, we allow P-GNNs to \textit{aggregate messages from anchor-sets}, which are randomly chosen subsets of all the nodes (Figure~\ref{fig:PGNN}, left). 
Note that anchor sets get resampled every time the model is run forward.
Secondly, when performing message aggregation, instead of letting each node aggregate information independently, the aggregation is \textit{coupled across all the nodes} in order to distinguish nodes with different positions in the network (Figure~\ref{fig:PGNN}, middle).
We design P-GNNs such that each node embedding dimension corresponds to messages computed with respect to one anchor-set, which makes the computed node embeddings position-aware (Figure~\ref{fig:PGNN}, right).


P-GNNs contain the following key components:\\
\-\hspace{5mm} $\bullet$ $k$ anchor-sets $S_i$ of different sizes.\\
\-\hspace{5mm} $\bullet$ Message computation function $F$ that combines feature information of two nodes with their network distance.\\
\-\hspace{5mm}$\bullet$ Matrix $\mb{M}$ of anchor-set messages, where each row $i$ is an anchor-set message $\mathcal{M}_i$ computed by $F$.\\
\-\hspace{5mm}$\bullet$ Trainable aggregation functions $\textsc{Agg}_M$, $\textsc{Agg}_S$ that aggregate/transform feature information of the nodes in the anchor-set and then also aggregate it across the anchor-sets.\\
\-\hspace{5mm}$\bullet$ Trainable vector $\mb{w}$ that projects message matrix $\mb{M}$ to a lower-dimensional embedding space $\mb{z} \in \mathbb{R}^k$.

Algorithm \ref{alg:pgnn} summarizes the general framework of P-GNNs.
A P-GNN consists of multiple P-GNN layers.
Concretely, the $l^\text{th}$ P-GNN layer first samples $k$ random anchor-sets $S_i$. Then, the $i^{\text{th}}$ dimension of the output node embedding $\mb{z}_{v}$ represents messages computed with respect to the $i^{\text{th}}$ anchor-set $S_i$.
Each dimension of the embedding is obtained by first computing the message from each node in the anchor-set via message computation function $F$, then applying a message aggregation function $\textsc{Agg}_M$, and finally applying a non-linear transformation to get a scalar via weights $\mb{w}\in\mathbb{R}^{r}$ and non-linearity $\sigma$.
Specifically, the message from each node includes distances that reveal node positions as well as feature-based information from input node features. The message aggregation functions are the same class of functions as used by existing GNNs. We further elaborate on the design choices in Section \ref{sc:design_choices}.

\xhdr{P-GNNs are position-aware}
The output embeddings $\mb{z}_v$ are position-aware, as each dimension of the embedding encodes the necessary information to distinguish structurally equivalent nodes that reside in different parts of the graph. Note that if we permute the dimensions of all the node embeddings $\mb{z}_v$, the resulting embeddings are equivalent to the original embeddings because they carry the same node positional information with respect to (permuted order of) anchor-sets $\{S_i\}$.

Multiple P-GNN layers can be naturally stacked to achieve higher expressive power.
Note that unlike GNNs, we cannot feed the output embeddings $\mb{z}_v$ from the previous layer to the next layer, because the dimensions of $\mb{z}_v$ can be arbitrarily permuted; therefore, applying a fixed non-linear transformation over this representation is problematic. The deeper reason we cannot feed $\mb{z}_v$ to the next layer is that the position of a node is always \textit{relative} to the chosen anchor-sets; thus, canonical position-aware embeddings do not exist.
Therefore, P-GNNs also compute structure-aware messages $\mb{h}_{v}$, which are computed via an order-invariant message aggregation function that aggregates messages \textit{across anchor-sets}, and are then fed into the next P-GNN layer as input. 



\begin{algorithm}[h!]
\caption{The framework of P-GNNs}
\label{alg:pgnn}
\begin{algorithmic}
\STATE {\bfseries Input:} Graph $G=(\mathcal{V},\mathcal{E})$; 
Set $S$ of $k$ anchor-sets $\{S_i\}$; 
Node input features $\{\mathbf{x}_v\}$;
Message computation function $F$ that outputs an $r$ dimensional message; 
Message aggregation functions $\textsc{Agg}_M, \textsc{Agg}_S$;
Trainable weight vector $\mb{w}\in\mathbb{R}^{r}$;
Non-linearity $\sigma$;
Layer $l \in [1,L]$
\STATE {\bfseries Output:} Position-aware embedding $\mb{z}_v$ for every node $v$
\STATE $\mathbf{h}_v \leftarrow \mathbf{x}_v$
\FOR{$l=1,\dots,L$} 
    \STATE $S_i \sim \mathcal{V}$\- for $i = 1,\dots,k$
    \FOR{$v \in \mathcal{V}$} 
        \STATE $\mb{M}_v = \mb{0} \in \mathbb{R}^{k\times r}$ 
        \FOR{$i = 1\dots,k$} 
            \STATE $\mathcal{M}_i \leftarrow \{F(v,u,\mb{h}_v,\mb{h}_u),
            \forall u \in S_i\}$ 
            \STATE $\mb{M}_v[i] \leftarrow \textsc{Agg}_M(\mathcal{M}_i)$ 
        \ENDFOR
        \STATE $\mb{z}_{v} \leftarrow \sigma(\mb{M}_v \cdot \mb{w})$
        \STATE $\mb{h}_{v} \leftarrow \textsc{Agg}_S(\{\mb{M}_v[i], \forall i \in [1,k]\})$
\ENDFOR
\ENDFOR   
 
\STATE $\mb{z}_v \in \mathbb{R}^k$, $\forall v \in \mathcal{V}$ 
\end{algorithmic}
\end{algorithm}

\subsection{Anchor-set Selection}
\label{sc:anchor_selection}
We rely on Bourgain's Theorem to guide the choice of anchor-sets, such that the resulting representations are guaranteed to have low distortion.
Specifically, distortion measures the faithfulness of the embeddings in preserving distances when mapping from one metric space to another metric space, which is defined as follows:
\begin{definition}
Given two metric spaces $(\mathcal{V},d)$ and $(\mathcal{Z},d')$ and a function $f: \mathcal{V} \rightarrow \mathcal{Z}$, $f$ is said to have distortion $\alpha$ if $\forall u,v \in \mathcal{V}$, $\frac{1}{\alpha} d(u,v) \leq d'(f(u),f(v)) \leq d(u,v)$.
\end{definition}

Theorem \ref{th:bourgain} states the Bourgain Theorem \cite{bourgain1985lipschitz}, which shows the existence of a low distortion embedding that maps from any metric space to the $l_p$ metric space:
\begin{theorem}
\label{th:bourgain}
(Bourgain theorem)
Given any finite metric space $(\mathcal{V},d)$ with $|\mathcal{V}| = n$, there exists an embedding of $(\mathcal{V}, d)$ into $\mathbb{R}^k$ under any $l_p$ metric, where $k = O(\log^2 n)$, and the distortion of the embedding is $O(\log n)$.
\end{theorem}

A constructive proof of Theorem \ref{th:bourgain} \cite{linial1995geometry} provides an algorithm to construct an $O(\log^2 n)$ dimensional embedding via anchor-sets, as summarized in Theorem \ref{th:bourgain_constructive}:

\begin{theorem}
\label{th:bourgain_constructive}
(Constructive proof of Bourgain theorem)
For metric space $(\mathcal{V},d)$, given $k = c\log^2 n$ random sets $S_{i,j} \subset \mathcal{V}, i=1,2,...,\log n, j = 1,2,...,c\log n$ where $c$ is a constant, $S_{i,j}$ is chosen by including each point in $\mathcal{V}$ independently with probability $\frac{1}{2^i}$. An embedding method for $v \in \mathcal{V}$ is defined as:
\begin{equation}
f(v) = \big( \frac{d(v, S_{1,1})}{k}, \frac{d(v, S_{1,2})}{k}, ..., \frac{d(v, S_{\log n,c\log n})}{k} \big)
\end{equation}
where $d(v, S_{i,j}) = \min_{u\in S_{i,j}} d(v,u)$. Then, $f$ is an embedding method that satisfies Theorem \ref{th:bourgain}.
\end{theorem}

The proposed P-GNNs can be viewed as a generalization of the embedding method in Theorem \ref{th:bourgain_constructive}, where the distance metric $d$ is generalized via message computation function $F$ and message aggregation function $\textsc{Agg}_M$ that accounts for both node feature information and position-based similarities (Section \ref{sc:design_choices}).
Using this analogy, Theorem \ref{th:bourgain_constructive} offers two insights for selecting anchor-sets in P-GNNs. First, $O(\log^2 n)$ anchor-sets are needed to guarantee low distortion embedding. Second, these anchor-sets have sizes distributed exponentially.
Here, we illustrate the intuition behind selecting anchor-sets with different sizes via the $1$-hop shortest path distance defined in Equation~\ref{eq:q_hop_dist}.
Suppose that the model is computing embeddings for node $v_i$. We say an anchor-set \emph{hits} node $v_i$ if $v_i$ or any of its one-hop neighbours is included in the anchor-set. Small anchor-sets can provide positional information with high certainty, because when a small anchor-set hits $v_i$, we know that $v_i$ is located close to one of the very few nodes in the small anchor-set. However, the probability that such small anchor-set hits $v_i$ is low, and the anchor-set is uninformative if it misses $v_i$. On the contrary, large anchor-sets have higher probability of hitting $v_i$, thus sampling large anchor-sets can result in high sample efficiency. However, knowing that a large anchor-set hits $v_i$ provides little information about its position, since $v_i$ might be close to any of the many nodes in the anchor-set.
Therefore, choosing anchor-sets of different sizes balances the trade-off and leads to efficient embeddings.

Following the above principle, P-GNNs choose $k = c\log^2 n$ random anchor-sets, denoted as $S_{i,j} \subset \mathcal{V}$, where $i=1,\dots,\log n, j = 1,\dots,c\log n$ and $c$ is a hyperparameter.
To sample an anchor-set $S_{i,j}$, we sample each node in $\mathcal{V}$ independently with probability $\frac{1}{2^i}$. 

\subsection{Design decisions for P-GNNs}
\label{sc:design_choices}

In this section, we discuss the design choices of the two key components of P-GNNs: the message computation function $F$ and the message aggregation functions $\textsc{Agg}$.

\xhdr{Message Computation Function $F$}
Message computation function $F(v,u,\mb{h}_v,\mb{h}_u)$ has to account for both position-based similarities as well as feature information. Position-based similarities are the key to reveal a node's positional information, while feature information may include other side information that is useful for the prediction task.

Position-based similarities can be computed via the shortest path distance, or, for example, personalized PageRank \cite{jeh2003scaling}. 
However, since the computation of shortest path distances has a $O(|\mathcal{V}|^3)$ computational complexity, we propose the following $q$-hop shortest path distance
\begin{equation}
  \label{eq:q_hop_dist}
    d^q_{sp}(v,u) = 
    \begin{cases}
    d_{sp}(v,u),  & \text{if } d_{sp}(v,u) \leq q, \\
    \infty, & \text{otherwise}
    \end{cases}
\end{equation}
where $d_{sp}$ is the shortest path distance between a pair of nodes.
Note that $1$-hop distance can be directly identified from the adjacency matrix, and thus no additional computation is needed. 
Since we aim to map nodes that are close in the network to similar embeddings, we further transform the distance $s(v,u) = \frac{1}{d^q_{sp}(v,u)+1}$ to map it to a $(0,1)$ range.


Feature information can be incorporated into $\mb{h}_u$ by passing in the information from the neighbouring nodes, as in GCN \cite{kipf2016semi}, or by concatenating node features $\mb{h}_v$ and $\mb{h}_u$, similar to GraphSAGE \cite{hamilton2017inductive}, although other approaches like attention can be used as well \cite{velickovic2017graph}.
Combining position and feature information can then be achieved via concatenation or product. We find that simple product works well empirically. Specifically, we find the following message passing function $F$ performs well empirically
\begin{equation}
   F(v,u,\mb{h}_v,\mb{h}_u) = s(v,u)  \textsc{concat}(\mb{h}_v,\mb{h}_u)
\end{equation}





\xhdr{Message Aggregation Functions $\textsc{Agg}$}
Message aggregation functions aggregate information from a set of messages (vectors).
Any permutation invariant function, such as $\textsc{Mean}, \textsc{Min}, \textsc{Max}, \textsc{Sum}$, can be used, and non-linear transformations are often applied before and/or after the aggregation to achieve higher expressive power \cite{zaheer2017deep}. 
We find that using simple $\textsc{Mean}$ aggregation function provides good results, thus we use it to instantiate both $\textsc{Agg}_M$ and $\textsc{Agg}_S$.






\section{Theoretical Analysis of P-GNNs}

\subsection{Connection to Existing GNNs}
\label{sc:connection_to_gnns}
P-GNNs generalize existing GNN models. From P-GNN's point of view, existing GNNs use the same anchor-set message aggregation techniques, but use different anchor-set selection and sampling strategies, and only output the structure-aware embeddings $\mb{h}_{v}$.

GNNs either use deterministic or stochastic neighbourhood aggregation \cite{hamilton2017inductive}.
Deterministic GNNs can be expressed as special cases of P-GNNs that treat each individual node as an anchor-set 
and aggregate messages based on $q$-hop distance. In particular, the function $F$ in Algorithm~\ref{alg:pgnn} corresponds to the message aggregation function of a deterministic GNN.
In each layer, most GNNs aggregate information from a node's one-hop neighbourhood \cite{kipf2016semi,velickovic2017graph}, corresponding to using $1$-hop distance to compute messages, or directly aggregating $k$-hop neighbourhood \cite{xu2018representation}, corresponding to computing messages within $k$-hop distance.
For example, a GCN \cite{kipf2016semi} can be written as choosing $\{S_i\} = \{v_i\}$, $\textsc{Agg}_M= \textsc{Mean}$, $\textsc{Agg}_S = \textsc{Mean}$, $F = \frac{1}{d^1_{sp}(v,u)+1}\mb{h}_u$, and the output embedding is $\mb{h}_u$ in the final layer.

Stochastic GNNs can be viewed as P-GNNs that sample size-1 anchor-sets, but each node's choice of anchor-sets is different. For example, GraphSAGE \cite{hamilton2017inductive} can be viewed as a special case of P-GNNs where each node samples $k$ size-1 anchor-sets and then computes messages using 1-hop shortest path distance anchor-set, followed by aggregation $\textsc{Agg}_S$. This understanding reveals the connection between stochastic GNNs and P-GNNs. First, P-GNN uses larger anchor-sets thereby enabling higher sample efficiency (Sec \ref{sc:anchor_selection}).
Second, anchor-sets that are shared across all nodes serve as reference points in the network, consequently, positional information of each node can be obtained from the shared anchor-sets.

\subsection{Expressive Power of P-GNNs}
\label{sc:anchor_distance}
Next, we show that P-GNNs provide a more \textit{general class of inductive bias} for graph representation learning than GNNs; therefore, are more expressive to learn both structure-aware and position-aware node embeddings.

We motivate our idea by considering pairwise relation prediction between nodes.
Suppose a pair of nodes $u, v$ are labeled with label $y$, using labeling function $d_y(u, v)$, and our goal is to predict $y$ for unseen node pairs.
From the perspective of representation learning, we can solve the problem via learning an embedding function $f$ that computes the node embedding $\mb{z}_v$, where the objective is to maximize the likelihood of the conditional distribution $p(y|\mb{z}_u, \mb{z}_v)$. Generally, an embedding function takes a given node $v$ and the graph $G$ as input and can be written as $\mb{z}_v = f(v, G)$, while $p(y|\mb{z}_u, \mb{z}_v)$ can be expressed as a function $d_z(\mb{z}_u, \mb{z}_v)$ in the embedding space.

As shown in Section \ref{sc:limitation_structure_aware}, GNNs instantiate $f$ via a function $f_{\theta}(v, S_v)$ that takes a node $v$ and its $q$-hop neighbourhood graph $S_v$ as arguments. Note that $S_v$ is independent from $S_u$ (the $q$-hop neighbourhood graph of node $u$) since knowing the neighbourhood graph structure of node $v$ provides no information on the neighbourhood structure of node $u$.
In contrast, P-GNNs assume a more general type of inductive bias, where $f$ is instantiated via $f_{\phi}(v, S)$ that aggregates messages from random anchor-sets $S$ that are shared across all the nodes, and nodes are differentiated based on their different distances to the anchor-sets $S$. 
Under this formulation, each node's embedding is computed similarly as in the stochastic GNN when combined with a proper $q$-hop distance computation (Section \ref{sc:connection_to_gnns}). However, since the anchor-sets $S$ are shared across all nodes, pairs of node embeddings are correlated via anchor-sets $S$, and are thus no longer independent.
This formulation implies a joint distribution $p(\mb{z}_u, \mb{z}_v)$ over node embeddings, where $\mb{z}_u = f_{\phi}(u, S)$ and $\mb{z}_v = f_{\phi}(v, S)$.
In summary, \textit{learning node representations} can be formalized with the following two types of objectives:

\-\hspace{5mm} $\bullet$ GNN representation learning objective:
\begin{equation}
\begin{aligned}
\label{eq:ob_edge_marginal}
    \min_\theta & \ \mathbb{E}_{u \sim V_{train}, v \sim V_{train}, S_u \sim p(V) , S_v \sim p(V)} \\
    & \mathcal{L}(d_z(f_{\theta}(u, S_u), f_{\theta}(v, S_v))-d_y(u,v))
\end{aligned}
\end{equation}
\-\hspace{5mm} $\bullet$ P-GNN representation learning objective:
\begin{equation}
\begin{aligned}
\label{eq:ob_edge_joint}
    \min_\theta & \ \mathbb{E}_{u \sim V_{train}, v \sim V_{train}, S \sim p(V)} \\
    & \mathcal{L}(d_z(f_{\phi}(u, S), f_{\phi}(v, S))-d_y(u,v))
\end{aligned}
\end{equation}
where $d_y$ is the target similarity metric determined by the learning task, for example, indicating links between nodes or membership to the same community, and $d_z$ is the similarity metric in the embedding space, usually the $l_p$ norm.

Optimizing Equations \ref{eq:ob_edge_marginal} and \ref{eq:ob_edge_joint} gives representations of nodes using joint and marginal distributions over node embeddings, respectively.
If we treat $u$, $v$ as random variables from $G$ that can take values of any pair of nodes, 
then the mutual information between the joint distribution of node embeddings and any $Y = d_y(u,v)$ is larger than that between the marginal distributions and $Y$: 
$I(Y; X_{joint}) \geq I(Y; X_{marginal})$, where 
$X_{joint} = (f_{\phi}(u, S_u), f_{\phi}(v, S_v)) \sim p(f_{\phi}(u, S_u), f_{\phi}(v, S_v))$;
$X_{marginal} = (f_{\theta}(u, S), f_{\theta}(v, S)) \sim p(f_{\theta}(u, S)) \otimes p(f_{\theta}(v, S))$, where $\otimes$ is the Kronecker product.
The gap of this mutual information is great, if the target task $d_y(u,v)$ is related to the positional information of nodes which can be captured by the shared choice of anchor-sets.
Thus, we conclude that P-GNNs, which embed nodes using the joint distribution of their distances to common anchors, have more expressive power than existing GNNs. 

\subsection{Complexity Analysis}
Here we discuss the complexity of neural network computation.
In P-GNNs, every node communicates with $O(\log^2 n)$ anchor-sets in a graph with $n$ nodes and $e$ edges. Suppose on average each anchor-set contains $m$ nodes, then there are $O(mn \log^2n)$ message communications in total. If we follow the exact anchor-set selection strategy, the complexity will be $O(n^2 \log^2n)$. In contrast, the number of communications is $O(n+e)$ for existing GNNs. 
In practice, we observe that the computation can be sped up by using a simplified aggregation $\textsc{Agg}_S$, while only slightly sacrificing predictive performance. 
Here for each anchor-set, we only aggregate message from the node closest to a given node $v$. This removes the factor $m$ in the complexity of P-GNNs, making the complexity $O(n\log^2n)$. 
We use this implementation in the experiments.

 \section{Experiments}

\begin{table*}[t]
\centering
\begin{footnotesize}
\caption{P-GNNs compared to GNNs on link prediction tasks, measured in ROC AUC. Grid-T and Communities-T refer to the transductive learning setting of Grid and Communities, where one-hot feature vectors are used as node attributes. Standard deviation errors are given.}
\label{tab:link_pred}
\begin{tabular}{@{}llllllllll@{}}
\toprule
               & Grid-T & Communities-T  &  Grid  & Communities & PPI   \\ \midrule
GCN             &$0.698 \pm 0.051$ &$0.981 \pm 0.004$ & $0.456 \pm 0.037$ & $0.512 \pm 0.008$  & $0.769 \pm 0.002$ \\
GraphSAGE        &$0.682 \pm 0.050$ &$0.978 \pm 0.003$ & $0.532 \pm 0.050$& $0.516 \pm  0.010$  & $0.803 \pm 0.005$ \\
GAT             &$0.704 \pm  0.050$ &$0.980 \pm 0.005$ & $0.566 \pm 0.052$& $0.618 \pm 0.025$  & $0.783 \pm 0.004$ \\
GIN              &$0.732 \pm 0.050$ &$0.984 \pm 0.005$ & $0.499 \pm 0.054$& $0.692  \pm 0.049$  & $0.782 \pm 0.010$ \\ \midrule
P-GNN-F-1L       &$0.542 \pm 0.057$ &$0.930 \pm 0.093$ & $0.619 \pm 0.080$& $0.939 \pm 0.083$  & $0.719 \pm 0.027$ \\
P-GNN-F-2L  &$0.637 \pm 0.078$ &$\mb{0.989} \pm 0.003$ & $0.694 \pm 0.066$& $\mb{0.991} \pm 0.003$  & $0.805 \pm 0.003$  \\\midrule
P-GNN-E-1L         &$0.665 \pm 0.033$ &$0.966 \pm 0.013$ & $0.879 \pm 0.039$& $0.985 \pm 0.005$  & $0.775 \pm 0.029$ \\
P-GNN-E-2L  &$\mb{0.834} \pm 0.099$ &$0.988 \pm 0.003$ & $\mb{0.940} \pm 0.027$& $0.985 \pm 0.008$  & $\mb{0.808} \pm 0.003$ \\ \bottomrule

\end{tabular}
\end{footnotesize}
\end{table*}

\begin{table}[t]
\centering
\begin{footnotesize}
\caption{Performance on pairwise node classification tasks, measured in ROC AUC. Standard deviation errors are given.}
\label{tab:community_detect}

\resizebox{\columnwidth}{!}{
\begin{tabular}{@{}llll@{}}
\toprule
               & Communities & Email & Protein  \\ \midrule
GAT              & $0.520 \pm 0.025$ & $0.515 \pm 0.019$ & $0.515 \pm 0.002$\\
GraphSAGE      & $0.514 \pm 0.028$ & $0.511 \pm 0.016$& $0.520 \pm 0.003$\\
GAT            & $0.620 \pm 0.022$ & $0.502 \pm 0.015$ & $0.528 \pm 0.011$\\
GIN             & $0.620 \pm 0.102$& $0.545 \pm 0.012$ & $0.523 \pm 0.002$\\ \midrule
P-GNN-F-1L      & $0.985 \pm 0.008$ & $0.630 \pm 0.019$ & $0.510 \pm 0.010$\\
P-GNN-F-2L & $0.997 \pm 0.006$ & $\mb{0.640} \pm 0.037$ & $\mb{0.729} \pm 0.176$\\\midrule
P-GNN-E-1L  & $0.991 \pm 0.013$ & $0.625 \pm 0.058$ & $0.507 \pm 0.006$\\
P-GNN-E-2L & $\mb{1.0} \pm 0.001$ & $\mb{0.640} \pm 0.029$ & $0.631 \pm 0.175$\\\bottomrule

\end{tabular}}
\end{footnotesize}
\end{table}




\subsection{Datasets}

We perform experiments on both synthetic and real datasets. We use the following datasets for a link prediction task:
\\
\-\hspace{5mm} $\bullet$ \xhdr{Grid
} 2D grid graph representing a 20$\times$ 20 grid with $|V|=$ 400 and no node features. 
\\
\-\hspace{5mm} $\bullet$ \xhdr{Communities}
Connected caveman graph \cite{watts1999networks} with 1\% edges randomly rewired, with 20 communities where each community has 20 nodes.
\\
\-\hspace{5mm} $\bullet$ \xhdr{PPI}
24 Protein-protein interaction networks \cite{zitnik2017predicting}. Each graph has 3000 nodes with avg. degree 28.8, each node has 50 dimensional feature vector.

We use the following datasets for pairwise node classification tasks which include community detection and role equivalence prediction\footnote{Inductive position-aware node classification is not well-defined due to permutation of labels in different graphs. However pairwise node classification, which only decides if nodes are of the same class, is well defined in the inductive setting.}.
\\
\-\hspace{5mm} $\bullet$ \xhdr{Communities}
The same as above-mentioned community dataset, with each node labeled with the community it belongs to.
\\
\-\hspace{5mm} $\bullet$ \xhdr{Emails}
7 real-world email communication graphs from SNAP \cite{leskovec2007graph} with no node features. Each graph has 6 communities and each node is labeled with the community it belongs to.
\\
\-\hspace{5mm} $\bullet$ \xhdr{Protein}
1113 protein graphs from \cite{borgwardt2005protein}. Each node is labeled with a functional role of the protein. Each node has a 29 dimensional feature vector.



\subsection{Experimental setup}



Next we evaluate P-GNN model on both transductive and inductive learning settings.

\xhdr{Transductive learning}
In the transductive learning setting, the model is trained and tested on a given graph with a fixed node ordering and has to be re-trained whenever the node ordering is changed or a new graph is given.
As a result, the model is allowed to augment node attributes with unique one-hot identifiers to differentiate different nodes.
Specifically, we follow the experimental setting from \cite{zhang2018link}, and use two sets of 10\% existing links and an equal number of nonexistent links as test and validation sets, with the remaining 80\% existing links and equal number of nonexistent links used as the training set.
We report the test set performance when the best performance on the validation set is achieved, and we report results over 10 runs with different random seeds and train/validation splits.

\xhdr{Inductive learning}
We demonstrate the inductive learning performance of P-GNNs on pairwise node classification tasks for which it is possible to transfer the positional information to a new unseen graph.
In particular, for inductive tasks, augmenting node attributes with one-hot identifiers restricts a model's generalization ability, because the model needs to generalize across scenarios where node identifiers can be arbitrarily permuted.
Therefore, when the dataset does not come with node attributes, we only consider using constant order-invariant node attributes, such as a constant scalar, in our experiments. Original node attributes are used if they are available.

We follow the transductive learning setting to sample links, but only use order-invariant attributes. When multiple graphs are available, we use 80\% of the graphs for training and the remaining graphs for testing. Note that we do not allow the model to observe ground-truth graphs at the training time. 
For the pairwise node classification task, 
we predict whether a pair of nodes belongs to the same community/class. In this case, a pair of nodes that do not belong to the same community are a negative example.

\subsection{Baseline models}
So far we have shown that P-GNNs are a family of models that differ from the existing GNN models.
Therefore, we compare variants of P-GNNs against most popular GNN models. To make a fair comparison, all models are set to have similar number of parameters and are trained for the same number of epochs. We fix model configurations across all the experiments. (Implementational details are provide in the Appendix.)
We show that even the simplest P-GNN models can significantly outperform GNN models in many tasks, and designing more expressive P-GNN models is an interesting venue for future work.

\xhdr{GNN variants}
We consider 4 variants of GNNs, each with three layers, including GCN \cite{kipf2016semi}, GraphSAGE \cite{hamilton2017inductive}, Graph Attention Networks (GAT) \cite{velickovic2017graph} and Graph Isomorphism Network (GIN) \cite{xu2018powerful}. Note that in the context of link prediction task, our implementation of GCN is equivalent to GAE \cite{kipf2016variational}.


\xhdr{P-GNN variants}
We consider 2 variants of P-GNNs, either with one layer or two layers (labeled 1L, 2L): (1) P-GNNs using truncated 2-hop shortest path distance (P-GNN-F); (2) P-GNNs using exact shortest path distance (P-GNN-E).



\subsection{Results}

\xhdr{Link prediction}
In link prediction tasks two nodes are generally more likely to form a link, if they are close together in the graph. Therefore, the task can largely benefit from position-aware embeddings.
Table \ref{tab:link_pred} summarizes the performance of P-GNNs and GNNs on a link prediction task.
We observe that P-GNNs significantly outperform GNNs across all datasets and variants of the link prediction taks (inductive vs. transductive). P-GNNs perform well in all inductive link prediction settings, for example improve AUC score by up to 66\% over the best GNN model in the grid dataset.
In the transductive setting, P-GNNs and GNNs achieve comparable performance.
The explanation is that one-hot encodings of nodes help GNNs to memorize node IDs and differentiate symmetric nodes, but at the cost of expensive computation over $O(n)$ dimensional input features and the failure of generalization to unobserved graphs. On the other hand, P-GNNs can discriminate symmetric nodes by their different distances to anchor-sets, and thus adding one-hot features does not help their performance.
In addition, we observe that when graphs come with rich features (e.g., PPI dataset), the performance gain of P-GNNs is smaller, because node features may already capture positional information. Quantifying how much of the positional information is already captured by the input node features is an interesting direction left for future work.
Finally, we show that the ``fast'' variant of the P-GNN model (P-GNN-F) that truncates expensive shotest distance computation at 2 still achieves comparable results in many datasets.

\xhdr{Pairwise node classification}
In pairwise node classification tasks, two nodes may belong to different communities but have similar neighbourhood structures, thus GNNs which focus on learning structure-aware embeddings will not perform well in this tasks.
Table \ref{tab:community_detect} summarizes the performance of P-GNNs and GNNs on pairwise node classification tasks. The capability of learning position-aware embeddings is crucial in the Communities dataset, where all P-GNN variants nearly perfectly detect memberships of nodes to communities, while the best GNN can only achieve 0.620 ROC AUC, which means that P-GNNs give 56\% relative improvement in ROC AUC over GNNs on this task. Similar significant performance gains are also observed in Email and Protein datasets: 18\% improvement in ROC AUC on Email and 39\% improvement of P-GNN over GNN on Protein dataset.

 \section{Conclusion}

We propose Position-aware Graph Neural Networks, a new class of Graph Neural Networks for computing node embeddings that incorporate node positional information, while retaining inductive capability and utilizing node features. We show that P-GNNs consistently outperform existing GNNs in a variety of tasks and datasets.

\section*{Acknowledgements}
\label{sec:ack}
This research has been supported in part by Stanford Data Science Initiative, NSF, DARPA, Boeing, Huawei, JD.com, and Chan Zuckerberg Biohub.

\bibliography{bibli}
\bibliographystyle{icml2019}

\end{document}